%% file: root.tex
\documentclass[letterpaper, 10 pt, conference]{ieeeconf}  

\IEEEoverridecommandlockouts                              

\usepackage{graphics,graphicx,float,subcaption,booktabs,xcolor,multirow,array,color,ifthen,tabu,colortbl,dblfloatfix,url,xparse,mathtools,patchcmd,amssymb,xspace,nicefrac,microtype,amsmath,amsthm,amsfonts,bm,ragged2e,tikz,stackengine,etoolbox,xpatch,enumerate,xstring,setspace,tabularx,makecell,changepage,cuted,titlesec,enumitem,wrapfig,tcolorbox,gensymb,algorithm,algpseudocode,makecell,cite}
\usepackage[bookmarksopen,bookmarksnumbered,
pdfpagemode=UseOutlines,
colorlinks=true,
linkcolor=red,
anchorcolor=blue,
citecolor=blue,
filecolor=blue,
menucolor=blue,
urlcolor=blue]{hyperref}
\usepackage[capitalise,noabbrev,nameinlink]{cleveref}
\usepackage[english]{babel}
\usepackage[font=small]{caption}

\newcommand{\website}{https://helpful-doggybot.github.io/}
\newcommand{\algo}{Helpful DoggyBot}

\usepackage{titlesec}
\titlespacing*{\section}{0pt}{0.2\baselineskip}{0\baselineskip}
\titlespacing*{\subsection}{0pt}{0.2\baselineskip}{0\baselineskip}
\titlespacing*{\paragraph}{0pt}{0\baselineskip}{0\baselineskip}

\title{\LARGE \bf
\algo: Open-World Object Fetching using\\Legged Robots and Vision-Language Models
}

\author{
Qi Wu\textsuperscript{1} \quad Zipeng Fu\textsuperscript{1} \quad Xuxin Cheng\textsuperscript{2} \quad Xiaolong Wang\textsuperscript{2} \quad Chelsea Finn\textsuperscript{1}\\
\textsuperscript{1}Stanford University \quad \textsuperscript{2}UC San Diego\\
\url{\website}
}

\begin{document}
\twocolumn[{%
\renewcommand\twocolumn[1][]{#1}%
\maketitle
\begin{center}
    \vspace{-0.3in}
    \centering
    \captionsetup{type=figure}
    \includegraphics[width=\textwidth]{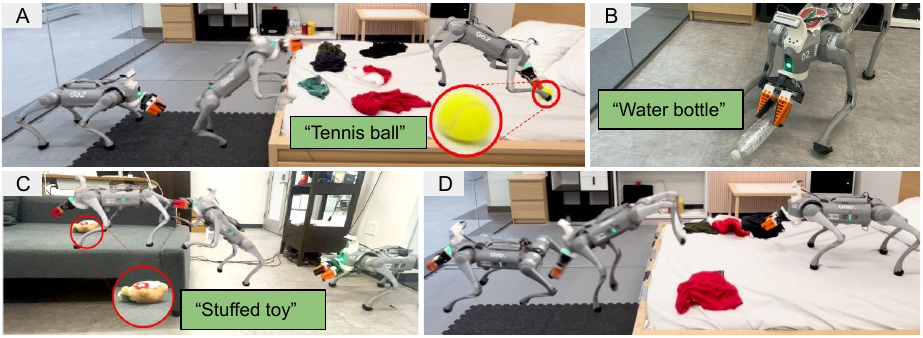}
    \captionof{figure}{\textbf{DoggyBot for Open-World Object Fetching}. Using the coordination of commands from VLMs and a low-level whole-body policy, our robot can (A) climb up a bed to fetch a tennis ball, (B) bend down to pick up a water bottle, (C) climb up a couch to pick up a stuffed toy, and (D) climb down the bed after retrieving the tennis bell.
    }
    \label{fig:teaser}
\end{center}%
}]

\thispagestyle{empty}
\pagestyle{empty}

\begin{abstract}
Learning-based methods have achieved strong performance for quadrupedal locomotion. However, several challenges prevent quadrupeds from learning helpful indoor skills that require interaction with environments and humans: lack of end-effectors for manipulation, limited semantic understanding using only simulation data, and low traversability and reachability in indoor environments. 
We present a system for quadrupedal mobile manipulation in indoor environments. It uses a front-mounted gripper for object manipulation, a low-level controller trained in simulation using egocentric depth for agile skills like climbing and whole-body tilting, and pre-trained vision-language models (VLMs) with a third-person fisheye and an egocentric RGB camera for semantic understanding and command generation.
We evaluate our system in two unseen environments without any real-world data collection or training. Our system can zero-shot generalize to these environments and complete tasks, like following user's commands to fetch a randomly placed stuff toy after climbing over a queen-sized bed, with a 60\% success rate. 
\end{abstract}

\section{Introduction}
Quadrupedal robots powered by learning-based methods have made significant strides in locomotion capabilities in recent years, demonstrating impressive agility and robustness across diverse terrains~\cite{rma,lee2020learning}.
However, their potential for assisting humans in everyday indoor environments remains largely untapped, 
like the ability to understand and follow language instructions to fetch a bottle of water for you. Several key challenges have hindered progress in this direction. First, equipping quadrupeds with effective manipulation capabilities without compromising agility is difficult, as traditional robotic arms often add significant weight and complexity~\cite{bellicoso2019alma}. While recent advances have demonstrated impressive quadrupedal agility, navigating cluttered indoor spaces and reaching high surfaces like beds or sofas requires a level of body control and environmental reasoning that goes beyond existing approaches. Moreover, bridging the semantic gap between simulation and reality remains a significant hurdle. Learning-based controllers trained in simulation often struggle to generalize to the rich, context-dependent nature of real-world indoor scenes, due to mismatches between simulation rendering and real-world sensing, and complexity in specifying diverse real-world scenarios in simulation. This limits robots' ability to understand and interact with diverse household objects and environments. 

In this paper, we present \algo{}, a quadrupedal robot system that aims to overcome these limitations and enable helpful mobile manipulation skills that can understand human commands and generalize across different indoor environments. To empower quadrupeds with general manipulation capabilities while still maintaining their agility, we design a simple yet effective 1-DoF gripper that is mounted on the bottom front of the robot. 
Shown in Figure~\ref{fig:teaser}, The gripper, serving as the ``mouth" of our robot, allows it to pick up and firmly hold everyday objects through ``biting". 

To increase the traversability and reachability of quadrupeds compared to prior work~\cite{agarwal2022legged,takahiro2022learning},
we use reinforcement learning and simulation to train a general-purpose low-level controller using egocentric depth and proprioception. We randomly sample robot commands including linear velocity, angular velocity, and pitch in a task-agnostic fashion in environments full of challenging obstacles during training. During zero-shot deployment in the real world after training, the controller takes in real-time egocentric depth measurements and task-specific commands. This controller equips robots with agile locomotion skills like climbing up a tall 0.5m obstacle and whole-body tilting while moving, powering quadrupeds to reach high workspaces like sofas
and beds and to grasp objects located on these places. These agile locomotion skills are also critical for navigating cluttered indoor spaces, where the ability to surmount various obstacles simplifies the otherwise complex maneuvers required for effective navigation. 

On the semantic perception and control front for solving useful tasks, instead of relying on collecting human demonstrations that is time-consuming or simulation that has semantic gaps, 
we leverage off-the-shelf VLMs to achieve zero-shot generalization in objects and configurations. Using VLMs and real-time video streams from a fish-eye top-down RGB camera mounted on the ceiling, our system can parse the open-vocabulary command of an object of interest, identify, localize and track the target object and robot itself within the scene, and generate reactive navigation commands based on the locations of the target object and the robot for the low-level controller. Upon approaching target objects, our system uses an egocentric RGB camera for tracking relative positions of the target object which are converted into velocity, pitch and grasping commands.

Our \algo{} integrates simulation training for low-level control, and VLMs for semantic understanding and command generation. We evaluate \algo{} in an unseen bedroom and an unseen living room, demonstrating its ability to complete open-vocabulary object fetching tasks like navigating to a randomly placed stuffed toy, climbing a bed and fetching back the toy from atop the bed with a 60\% success rate. Notably, our system achieves this generalization without any real-world data collection or training, highlighting the potential of our approach for creating helpful quadrupedal assistants that can adapt to diverse home environments. The key contributions of our system include (1) a simple yet effective 1-DoF gripper design that enables object grasping for quadrupeds, 
(2) a general-purpose low-level controller trained in simulation that enables real-world parkour-like mobility, (3) an approach leveraging pre-trained VLMs for semantic understanding for quadrupeds through generating reactive velocity, pitch and grasping commands,
and (4) experimental validation demonstrating zero-shot generalization to unseen indoor configurations for mobile manipulation tasks.

\section{Related Work}
\textbf{Legged Mobile Manipulation.} Legged robots have long been of interest for their potential to traverse complex terrains while performing manipulation tasks. Much work in this area focused on bipedal humanoid platforms~\cite{kato1973development,hirai1998development,chignoli2021humanoid,nelson2012petman,stasse2017talos,radford2015valkyrie},
demonstrating basic manipulation while maintaining balance~\cite{harada2005humanoid,arisumi2007dynamic,settimi2016motion,ferrari2017humanoid,harada2007real,sato2021drop,fu2024humanplus,cheng2024tv,he2024omnih2o,dao2023sim}. More recently, quadrupedal robots have gained attention for their inherent stability and agility~\cite{park2015online,nguyen2019optimized,nguyen2022continuous,gehring2016practice,smith2021legged,ma2024dreureka,margolis2022rapid,ji2022concurrent,hwangbo2019learning,yang2023continuous,cheng2023parkour,zhuang2023robot,rudin2022advanced,yang2022cerberus,agarwal2022legged,takahiro2022learning,rma,kang2023rl}. Several approaches have been explored to enable manipulation capabilities on quadrupeds. One common method is to mount a robotic arm on the quadruped's back~\cite{ferrolho2023roloma,ma2022mani,zimmermann2021go,bellicoso2019alma,liu2024visual,portela2023learning,ha2024umilegs,fu2022deep,yokoyama2023adaptive,pan2024roboduet}. While this provides significant dexterity, it also adds considerable weight and complexity to the system, hence reducing the agility of quadrupeds. An alternative approach is to utilize the quadruped's existing or modified limbs and torso for simple pushing tasks~\cite{jeon2023learning,lin2024locoman,cheng2023legs,he2024learning,xu2023creative,arm2024pedipulate}. Learning-based methods have shown promise in developing legged manipulation skills. For instance, \cite{ha2024umilegs}
combines imitation learning for target end-effector trajectory generation and reinforcement learning for low-level control. \cite{cheng2023legs} chains multiple polices to complete pushing tasks guided by fiducial markers. However, these approaches often struggle to generalize beyond the specific tasks and environments used during training. Our work builds upon these foundations by introducing a simple yet effective gripper design and a learning approach that enables generalization to unseen environments. Unlike previous work, we focus on enabling helpful indoor tasks that require both agile locomotion and object manipulation.

\textbf{Robot Learning using Large Pretrained Models.} The advent of large pre-trained models, particularly in the domains of computer vision and natural language processing, has opened new avenues for robot learning~\cite{open_x_embodiment_rt_x_2023,huang2023voxposer,driess2023palme,yu2023language,yuan2024robopoint,nasiriany2024pivot,chen2024commonsense,huang2024copa,liu2024moka,du2023video,chen2024spatialvlm,gao2024physically,duan2024manipulate,rt22023arxiv,shridhar2021cliport,liang2023code,kim2024openvla,zhen20243d,black2023zero,moo2023arxiv}.
These models, trained on vast amounts of visual data, offer rich semantic representations that can be leveraged for various robotic tasks. In the context of manipulation, \cite{huang2023voxposer,yu2023language} use VLMs to generate cost functions for tasks specified in language instructions, while \cite{nasiriany2024pivot,gu2023rt,sundaresan2024rt,rth2024arxiv} use VLMs to directly generate executable commands or intermediate presentations.
These approaches, however, were primarily focused on static manipulation scenarios. For mobile robots, recent work has explored using large pretrained models for navigation and locomotion~\cite{shah2022lmnav,xu2023creative,wu2023tidybot,tang2023saytap,nasiriany2024pivot,chen2024identifying}. 
For example, prior work demonstrates how VLMs can be used to generate navigation commands for wheeled robots~\cite{shah2022lmnav} and legged robots~\cite{chen2024commonsense}. However, the integration of these models with mobile manipulation remains relatively unexplored. Our work bridges this gap by leveraging pretrained vision-language models to enable semantic understanding and adaptive behavior generation for a quadrupedal robot performing mobile manipulation tasks. Unlike previous approaches, we demonstrate how large pretrained models can be effectively used in conjunction with learned low-level controllers to enable zero-shot generalization to mobile manipulation tasks.

\section{Hardware}
\begin{figure}
    \centering
    \includegraphics[width=0.8\linewidth]{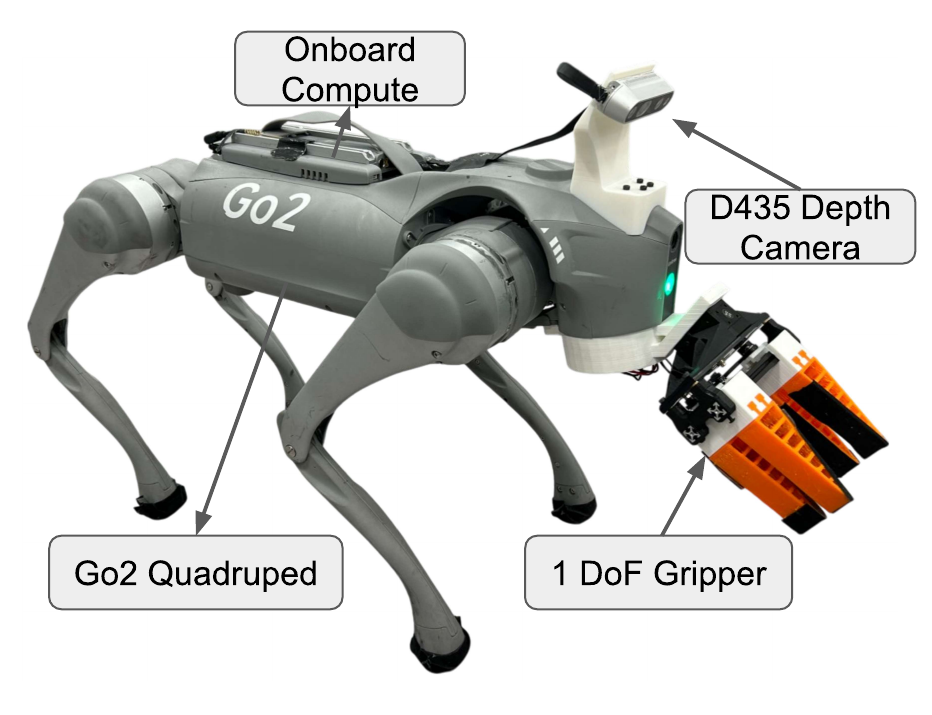}
    \caption{\textbf{Hardware Setup.} We use a Go2 quadruped and a custom-built 3D printed Gripper actuated by Dynamixel XM430-W350-T servo motor. An egocentric RealSense D435 is mounted on the top front of the robot with 30 degrees downwards.}
    \label{fig:hardware}
\end{figure}
Shown in Figure~\ref{fig:hardware}, our robot hardware system consists of a 12-DoF Unitree Go2 quadruped robot and a 1-DoF gripper mounted on the bottom from of the robot. Both are powered by the onboard battery of Go2. We 3D-print and custom-build our Finray gripper which is actuated by a Dynamixel XM430-W350-T servo motor through a slider-crank mechanism for fast closing. We use the onboard Jetson to run our learned low-level controller that takes egocentric depth from a RealSense D435 and proprioception as input, and VLMs upon approaching objects that takes egocentric RGB as input. We send high-level commands, that are generated from VLMs running on a separate workstation that takes third-person top-down RGB stream as input, to the Jetsen through Wi-Fi.

\section{Learning a General Whole-Body Controller}
\begin{figure*}[t]
    \centering
    \includegraphics[width=0.9\textwidth]{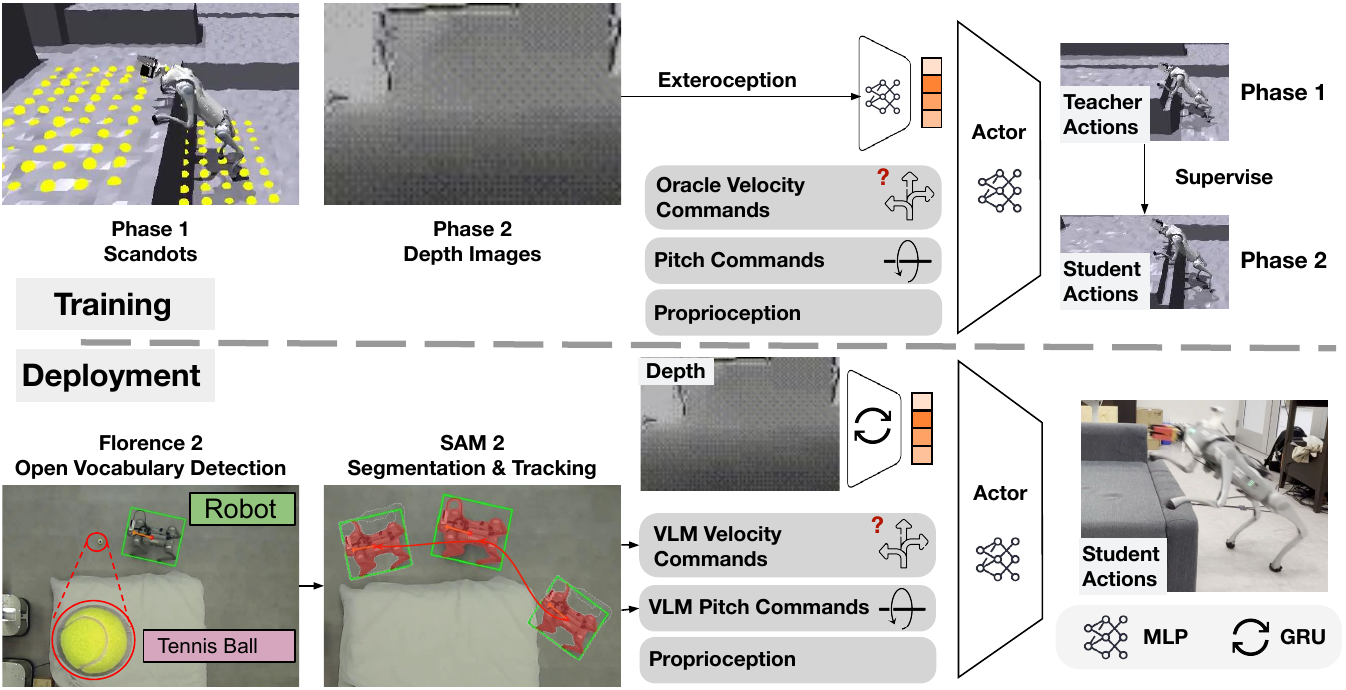}
    \caption{\textbf{System Overview.} We use a two-phase framework to train a depth-based policy as the low-evel whole-body controller. During deployment, we use VLMs for open-vocabulary detection, segmentation and tracking models to provide velocity commands and pitch commands for the controller.}
    \label{fig:method}
\end{figure*}

To enable effective mobile manipulation in diverse indoor environments, our robot requires both agile locomotion skills for traversing challenging obstacles and precise whole-body control for expanding its workspace. Previous works have typically addressed these challenges separately~\cite{cheng2023parkour,zhuang2023robot}, but integrating multiple objectives into a single learning framework introduces new complexities. These include increased exploration burden and the potential for sub-optimal behaviors when optimizing for multiple objectives simultaneously~\cite{fu2022deep}. Shown in Figure~\ref{fig:method}, our approach leverages a two-phase training process focusing on the whole-body control and agility to overcome these challenges.
\subsection{Phase 1: Training with Privileged Information}
We develop our agile visual whole-body control policy through a two-phase training process:
In the first phase, we train a policy using PPO~\cite{schulman2017proximal} to optimize both whole-body control and agile locomotion objectives. During this phase, the policy uses privileged information in the form of scandots, capturing heights of terrain near the robot, as observations, allowing for efficient learning in simulation.

\textbf{Whole-body objective:} This objective enables the robot to track a randomly sampled pitch command, expanding the workspace of the 1-DoF gripper. We define the reward as:
\begin{equation}
r_{\text{wb}} = \exp(-3 \cdot |p_{\text{cmd}} - p|)
\end{equation}
where $p_{\text{cmd}}$ is the commanded pitch uniformly sampled from the range $[-30^\circ, 30^\circ]$ and $p$ is the actual pitch angle of the robot's body. We remove this objective only when the robot is encourtering obstacles to avoid conflicting objectives.

\textbf{Agile locomotion objective:} This objective encourages the robot to traverse challenging obstacles such as high steps. To mitigate the exploration burden, we adopt a velocity tracking reward inspired by~\cite{cheng2023parkour}:
\begin{equation}
r_{\text{tracking}} = \min(\langle v,\hat{d}_{\text{wp}} \rangle, v_{\text{cmd}}) / v_{\text{cmd}}
\label{eq:inner_prod}
\end{equation}
where $v\in\mathbb{R}^2$ is the robot's current velocity in the world frame, $v_{\text{cmd}}\in\mathbb{R}$ is the linear velocity command sampled from the range [0, 1m/s], and $\hat{d}_\text{wp}$ is the unit vector pointing towards the next waypoint. We convert $\hat{d}_{\text{wp}}$ into angular velocity command $\omega_{\text{cmd}}$ as a policy input, which calculates the angular difference between robot's current direction and $\hat{d}_\text{wp}$, removing dependency of the policy on global information. We track velocity in the world frame to prevent the robot from learning unintended behaviors like circumventing obstacles.
We compute the direction using waypoints placed on the terrain:
\begin{equation}
\hat{d}_{\text{wp}} = \frac{x_\text{wp} - x}{|x_\text{wp} - x|}
\end{equation}
where $x_w$ is the location of next waypoint and $x$ is the robot's current position in the world frame.

\subsection{Phase 2: Policy Distillation using Egocentric Depth}
To enable real-world deployment, we distill the learned policy during Phase 1 into a deployable policy that operates on depth images from a front-facing camera instead of privileged scandots information. We use Regularized Online Adaptation (ROA)~\cite{fu2022deep} to train an online estimator to recover environmental information from the
history of onboard observations.
Our online estimator architecture consists of a convolutional neural network (CNN) followed by a gated recurrent unit (GRU) to process the temporal sequence of depth images. This design allows the policy to capture both spatial and temporal information from the visual input. The output of this estimator replaces the scandots input to the base policy learned in Phase 1.
A key difference from previous work~\cite{cheng2023parkour} is that we do not perform dual distillation of both the heading command and the exteroception simultaneously. Instead, we leverage a more powerful VLM to specify the robot's intended heading direction. This approach helps us avoid potential out-of-distribution problems that can arise in dual-distillation processes.

\subsection{Simulation Environments and Training Curricula}
To ensure robust performance across diverse scenarios, we train our policy in a variety of simulated environments featuring challenging obstacles such as stairs and uneven terrain. We randomly generate these environments for each training episode, varying parameters like stair height, number of stairs, and terrain friction to promote generalization.

To further improve learning efficiency and policy performance, we employ reward shaping techniques and a curriculum learning approach. We introduce auxiliary rewards for maintaining balance, minimizing energy consumption, and smooth transitions between different locomotion modes (e.g., walking, climbing, and tilting). The curriculum progressively increases the difficulty of the training environments, starting with simple flat terrains and gradually introducing more complex obstacles as the policy improves.
By combining these techniques with our two-phase training process, we develop a general whole-body controller capable of agile locomotion and precise manipulation in diverse indoor environments. This approach enables our quadrupedal robot to navigate challenging obstacles and perform complex mobile manipulation tasks without requiring extensive real-world data collection or environment-specific training. More details can be found on the \href{\website}{project website}.

\section{Zero-Shot Deployment using VLMs}

\begin{table*}[t]
    \centering
    \begin{tabular}{l|ccccc|ccccc}
        \toprule
        & \multicolumn{5}{c|}{Success Rate (\%) $\uparrow$} & \multicolumn{5}{c}{Average Distance (\%) $\uparrow$} \\
        & \thead{Climb\\Up} & \thead{Climb\\Down} & \thead{Walk\\$30^{\circ}$ pitch} & \thead{\thead{Walk\\$-30^{\circ}$ pitch}} & \thead{Walk} & \thead{Climb\\Up} & \thead{Climb\\Down} & \thead{Walk\\$30^{\circ}$ pitch} & \thead{\thead{Walk\\$-30^{\circ}$ pitch}} & \thead{Walk} \\
        \midrule
        Blind & 0 & 0 & 100 & 100 & 100 & 11 & 10 & 100 & 100 & 100 \\
        No GRU & 0 & 0 & 100 & 100 & 100 & 12 & 13 & 100 & 100 & 100 \\
        No Distill & 0 & 0 & 0 & 0 & 0 & 0 & 0 & 0 & 0 & 0 \\
        No Waypoint & 14 & 12 & 90 & 100 & 100 & 10 & 13 & 100 & 100 & 100 \\
        Ours & \textbf{96} & \textbf{90} & 100 & 100 & 100 & \textbf{92} & \textbf{84} & 100 & 100 & 100 \\
        \midrule
        Oracle (Phase 1) & 98 & 96 & 100 & 100 & 100 & 95 & 92 & 100 & 100 & 100 \\
        \bottomrule
    \end{tabular}
\caption{\textbf{Simulation Results.} We compare our learned controller with five baselines in simulation. Our performs the best in all tasks in both success rate and average distance reached. We find only small degradation in performance from the oracle policy using priviledged information in Phase 1. All metrics are averaged across 200 trials.}
\label{tab:sim-baselines}
\end{table*}

To enable zero-shot generalization to unseen environments and objects, we leverage pre-trained VLMs for semantic understanding and adaptive behavior generation. Our system integrates open-vocabulary object detection, efficient navigation, and precise grasping, all without requiring task-specific training data or fine-tuning.

\subsection{Open-Vocabulary Detection, Segmentation and Tracking}

Our system employs a combination of state-of-the-art vision models to achieve robust open-vocabulary object detection, segmentation and tracking.

\textbf{Initial Detection}: We utilize Florence-2~\cite{xiao2024florence} to perform open-vocabulary object detection. This allows our system to identify and localize both the robot itself and the target objects based on natural language descriptions, enabling flexibility in task specification. 

\textbf{Segmentation}: Following initial detection, we apply SAM2 (Segment Anything Model 2)~\cite{ravi2024sam} to generate precise object masks. The integration of Florence-2 and SAM2 enables our system to handle a wide range of objects without prior training on specific categories.

\textbf{Tracking}: To maintain real-time performance, we employ SAM2 for object tracking at 10 Hz. This approach allows for continuous updating of the object's position in the environment, crucial for navigation and manipulation tasks.

\subsection{Navigation}
Our navigation system leverages a top-down fisheye camera mounted on the ceiling to provide a global view of the environment. This perspective enables simultaneous tracking of both the robot and target object positions, simplifying the planning process. We use the detected object position as a single waypoint for navigation, generating commands that guide the robot efficiently towards its goal. The system maintains a constant linear velocity of 0.8 m/s towards the waypoint, while angular velocity is computed using a proportional controller with $K_p$ = 0.5 based on the difference between the robot's current heading and the vector pointing to the waypoint. During this phase, the pitch command is set to 0. To ensure smooth integration of locomotion and manipulation, the system transitions from navigation to grasping mode when the robot is approximately 1 meter away from the target object. We assume that our low-level controller can traverse most indoor obstacles like beds and sofas, thus alleviating the need for obstacle avoidance. 

\subsection{Grasping Objects}
As the robot approaches the target object, it switches to a precise grasping strategy using its front-mounted gripper, transitioning from global to egocentric perception. The system now relies on egocentric depth and RGB cameras mounted on the robot for fine-grained control. Since SAM2 is compute-intensive and hence unsuitable for onboard inference, we employ an on-device multi-stage perception pipeline for accurate object localization, combining GroundingDINO~\cite{liu2023grounding} for object detection at 0.2 Hz, MobileSAM~\cite{zhang2023mobile} for generating precise object masks on the RGBD input at 0.2 Hz, and Cutie~\cite{cheng2023putting} for high-frequency tracking at 10 Hz. This approach maintains accurate object position information between slower detection updates. From the tracked mask, we extract the (x, y, z) coordinates of the object's center in the robot's local frame. The grasping commands are then generated using proportional controllers: the linear velocity command is controlled based on the x-coordinate with $K_p$ = 0.5, the angular velocity command is adjusted using the y-coordinate with $K_p$ = 0.5, and the pitch command is computed from the z-coordinate with $K_p$ = 1. The system triggers the grasping action when all coordinates are within a small threshold, indicating optimal positioning relative to the target object.

By integrating these components, our system achieves zero-shot generalization to new configurations and objects, enabling the quadrupedal robot to perform complex mobile manipulation tasks without environment-specific training or data collection. The use of pre-trained VLMs and efficient perception pipelines allows for robust performance across a wide range of scenarios, making our approach suitable for diverse indoor applications.

\begin{table*}[t]
    \centering
    \setlength\tabcolsep{5pt}
    \begin{tabular}{l|cccc|cccc|ccc|ccc}
        \toprule
        & \multicolumn{11}{c|}{First-Attempt Success Rate (\%) $\uparrow$} & \multicolumn{3}{c}{Average Time (s) $\downarrow$} \\
        & \multicolumn{4}{c|}{\small Bed + Toy} & 
        \multicolumn{4}{c|}{\small Sofa + Bottle} &
        \multicolumn{3}{c|}{\small Ground + Ball} &
        \multirow{3}{*}{\thead{Bed +\\Toy}} & 
        \multirow{3}{*}{\thead{Sofa +\\Bottle}} &
        \multirow{3}{*}{\thead{Ground + \\Ball}} \\
        & \thead{navigate +\\climb up} & \thead{pick\\up} & \thead{climb\\down} & \thead{\textit{Total}} & 
        \thead{navigate +\\climb up} & \thead{pick\\up} & \thead{climb\\down} & \thead{\textit{Total}} & \thead{navigate} & \thead{pick\\up} & \thead{\textit{Total}}  \\
        \midrule
        Go2 Default & 0 & 0 & 0 & 0 & 0 & 0 & 0 & 0 & 80 & 0 & 0 & - & - & -\\
        No Tracking & 60 & 0 & 0 & 0 & 50 & 0 & 0 & 0 & 40 & 0 & 0 & – & - & - \\
        Ours & 90 & 78 & 86 & \textbf{60} & 80 & 88 & 86 & \textbf{60} & 100 & 70 & \textbf{70} & \textbf{62} & \textbf{50} & \textbf{23} \\
        \midrule
        Teleop & 90 & 89 & 100 & 80 & 90 & 89 & 88 & 70 & 100 & 80 & 80 & 75 & 58 & 38 \\
        \bottomrule
    \end{tabular}
\caption{\textbf{Real-World Results.} We compare our system with three baselines including Go2 default controller instead of our learned controller, teleoperation using a remote controller instead of using VLMs, and ours without reactive tracking powered by SAM2. Ours outperform all baselines in average time to completion, and close to teleoperation in success rates. We measure the success rates and average time to completion across 10 trials per setting.}
\label{tab:real-baselines}
\end{table*}

\section{Experiments}
\subsection{Simulation Experiments}
\textbf{Baselines in Simulation.} We compare our controller with several baselines including \textit{Blind}, \textit{No GRU}, \textit{No Distill} and \textit{No Waypoint}. We also include \textit{Oracle (Phase 1)}, the policy trained in phase 1 using privileged information.
\begin{itemize}[noitemsep,leftmargin=1.2em,itemsep=0em,topsep=0em]
    \item \textbf{Blind}: a blind policy using only proprioception and no depth images as observations. 
    \item \textbf{No GRU}: a MLP policy baseline. Instead of using a GRU, it uses only the depth image and proprioception at the current time step without any memory to predict actions. 
    \item \textbf{No Distill}: an ablation training a deployable policy with GRU directly using PPO with our two-phase training process, so skipping the distillation stage.
    \item \textbf{No Waypoint}: removing the agile locomotion objective guided by waypoints. Directly train the policy in Phase~1 with a reward encouraging tracking sampled linear and angular velocity commands.
    \item \textbf{Oracle (Phase 1)}: The policy from the first training phase, which has access to privileged information only available in simulation, such as terrain scandots.
\end{itemize}
These baselines allow us to assess the impact of various components in our approach, including the importance of visual input, temporal memory, the two-phase training process, and the use of waypoints in guiding robot forward. Additionally, comparing against the Oracle provides insight into the performance gap between our deployable policy and one with access to privileged environmental information.

\textbf{Simulation Results.}
Shown in Table~\ref{tab:sim-baselines}, the Blind and No GRU baselines exhibit poor performance, failing in most tasks except for the simple Walk task where they achieve 100\% success. These baselines lack the necessary spatial awareness or memory mechanisms required for complex sequential navigation tasks involving climbing. Learning from vision directly increases the complexity of the training process, where the network can't learn from scratch properly. The No Waypoint baseline shows moderate success in Climb Up and Climb Down, but still struggles with the more challenging climbing tasks, highlighting the importance of on-the-fly velocity command generation for climbing. Without waypoints as guidance, the robot easily learns to walk pass the obstacle or turn around instead of trying to climb as a result of rewarding local velocity only. In contrast, our approach achieves consistently higher performance, with near-perfect scores in most tasks, especially Climb Up and Climb Down, and outperforms all baselines. We find only small degradation in performance from the oracle policy using priviledged information in Phase 1. Our approach shows that distilled policy can perform as well as the Oracle policy, suggesting the effectiveness of the two-phase training process. The overall results demonstrate the importance of integrating components such as depth information, memory, waypoint guidance and distillation.

\subsection{Real-World Experiments}
\textbf{Baselines and Tasks in Real World.} We compare our system deployed in the real world with several baselines. The baselines include \textit{Go2 Default}, \textit{Teleop}, and \textit{No Tracking}: 
\begin{itemize}[noitemsep,leftmargin=1em,itemsep=0em,topsep=0em]
    \item \textbf{Go2 Default}: the default controller built in with Go2. This controller does not use exteroception.
    \item \textbf{Teleop}: the commands are generated by an expert human operator through a remote controller, replacing VLMs.
    \item \textbf{No Tracking}: the commands are generated open-loop using the initial pose detection of the robot itself and the object of interest. 
\end{itemize}
Illustrated in Figure~\ref{fig:teaser}, we select three objects and three environments that represent realistic real-world scenarios:
\begin{itemize}[noitemsep,leftmargin=1.2em,itemsep=0em,topsep=0em]
    \item \textbf{Bed + Toy}: The robot needs to fetch a stuffed toy on a bed. The task requires the robot to climb up a queen-sized bed with 40cm in height, pick up the stuffed toy on the bed, and climb down the bed. The stuffed toy is placed uniformly randomly on a 1m by 1m region on the bed. The robot is initially randomly placed in the bedroom.
    \item \textbf{Sofa + Bottle}: The robot needs to fetch an empty plastic water bottle on a sofa. The task requires the robot to climb up a sofa with a height of 44cm, pick up the bottle on the sofa, and climb down the sofa. The bottle is placed uniformly randomly on a 0.2m by 1m region on the sofa. The robot is initially randomly placed in the room.
    \item \textbf{Ground + Ball}: The robot needs to fetch a ball on the ground. The ball is placed uniformly randomly on a 3m by 3m region on the ground. 
\end{itemize}
We test our system and the three baselines on all four tasks. We measure the success rates and average time to completion across 10 trials per setting. Qualitative results can be found on the \href{\website}{project website}.

\textbf{Real-World Results.} 
The real-world experiments, as summarized in Table~\ref{tab:real-baselines}, demonstrate the effectiveness of our system compared to the three baselines. In the task involving navigating to a toy on a bed, our system achieved a 60\% total first-attempt success rate, significantly outperforming the Go2 default controller and No Tracking baselines, both of which failed to complete the task. Go2 default controller fails to climb up high obstacles like beds and sofas, whereas No Tracking only generates an open-loop trajectory of commands and fails to compensate drifting in navigation and subsequent grasping. Our system's performance was close to that of teleoperation, with only a 20\% gap in the first-attempt success rate. We find that though teleoperation can solve tasks perfectly given many attempts, the first-attempt success rates of teleoperation are only around 70-80\% given an expert human operator. Similarly, in the task of fetching a bottle from a sofa with soft deformabile, our approach achieves a 60\% success rate, close to teleportation. This tasks also demonstrates the robustness of our learned controller in walking on soft deformable surfaces. In the Ground + Ball task, which involves simpler navigation and grasping on flat terrain, our system achieving a 70\% success rate, outperforming all baselines than teleoperation. In terms of average time to completion, our system consistently outperformed the baselines, completing tasks faster than both the Go2 Default and No Tracking methods. Notably, our system was also faster than teleoperation, particularly in the Ground + Ball task, where it completed the task in 23 seconds on average compared to teleoperation’s 38 seconds. These results highlight the strength of our approach in achieving open-vocabulary object fetching in novel environments under a reasonable amount of time.

\section{Conclusion, Limitations \& Future Directions}
We presented \algo{}, a quadrupedal robot system capable of zero-shot mobile manipulation in diverse indoor environments, integrating a 1-DoF gripper, learned whole-body control, and vision-language models. While our approach demonstrates progress, limitations include the gripper's restricted dexterity, reliance on ceiling-mounted cameras for navigation, and potential occlusion to the perception system. In future work, we will focus on enhancing manipulation capabilities without compromising agility, developing navigation strategies using only onboard sensors, and future improving agility to achieve cheerful pet behaviors~\cite{duan2024playful}. Additional places for improvement include integrating multiple tasks into complex sequences, improving robustness in dynamic environments, incorporating online learning and human feedback, and exploring societal implications of advanced quadrupedal robots in domestic settings. By addressing these challenges, this research direction has the potential to revolutionize human-robot interaction and assistance in daily life.

\section*{Acknowledgement}
We thank the hardware and firmware supports from Unitree Robotics. We appreciate the initial brainstorming, valuable discussions and constructive feedback from Ziwen Zhuang and Xin Duan. We appreciate long-term supports on hardware and code from and discussions with Huy Ha and Yihuai Gao. We appreciate the help in experiments from Ziang Cao, Tian-Ao Ren and Hang Dong. We also appreciate discussions with Wenhao Yu and Erwin Coumans. This project is supported by the AI Institute and ONR grant N00014-21-1-2685. Zipeng Fu is supported by Pierre and Christine Lamond Fellowship.

\clearpage
\bibliographystyle{IEEEtran}
\bibliography{root}

\clearpage
\appendix
\input{supplementary-text}

\end{document}

%% file: supplementary-text.tex
\subsection{Details of Whole-Body Controller}
\subsubsection{Details of Simulation Environment}
To ensure robust performance across diverse scenarios for the expert policy, we use Isaac Gym Preview 4 to train 6144 robots in 400 terrains. We introduce a curriculum learning approach which generates 10 different levels of stair heights in simulation. The criteria of updating the curriculum in training is the proportion of the terrain each episode the robot finishes. We randomly generate these environments for each training episode, varying parameters like stair height, number of stairs, and terrain friction as shown in Table~\ref{tab:envsettings}. We then distilled a policy with 384 robots in simulation with real-time depth image rendering. For Oracle policy, we trained 20k iterations in 10 hours on a single GeForce RTX 4090 GPU. For distilled policy, we trained 5k iterations in 6 hours.

\begin{table}[ht]
    \centering
    \begin{tabular}{c|c}
        \toprule
        Parameters & Values \\
        \midrule
        num of envs & 6144 \\
        num of vision envs & 384 \\
        \midrule
        num of terrains & 400 \\
        num of difficulty levels & 10 \\
        stair height & [0, 0.65] \\
        stair per env & [0, 6] \\
        stair width & [0.8, 3]\\
        stair length & [1.5, 2] \\ 
        goal y range & [-0.1,0.1] \\
        terrain noise & [0.02, 0.06]  \\
        friction & [0.2, 2] \\
        curriculum up threshold & 0.8*total length \\
        curriculum down threshold & 0.5*total length \\
        \bottomrule
    \end{tabular}
\caption{\textbf{Environment and terrain setup}}
\label{tab:envsettings}
\end{table}

\subsubsection{Details of Domain Randomization}
To further promote generalization and ensure robust performance in real world application, we employ domain randomization. We uniformly sample values in Table~\ref{tab:domain_rand} to change the robots' dynamics and perturbations, enabling it to bridge the sim2real gap.

\begin{table}[ht]
    \centering
    \begin{tabular}{c|c}
        \toprule
        Parameters & Values \\
        \midrule
        push interval (s) & 8 \\
        max push vel xy (m/s) & 0.5 \\
        max push vel z (m/s) & 0.5 \\
        added mass range (kg) & [0., 3.] \\
        added com range (m) & [-0.2, 0.2] \\
        motor strength range & [0.8, 1.2] \\
        action delay(s) & [0 0.02] \\
        \midrule
        vision delay(s) & 0.1 \\ 
        vision position rand(m) & 0.005 \\
        vision angle rand(degree) & [24,34] \\
        \bottomrule
    \end{tabular}
\caption{\textbf{Domain randomization}}
\label{tab:domain_rand}
\end{table}

\subsubsection{Details of Reward Function}
To enhance learning efficiency and policy performance, we employed reward shaping techniques. Auxiliary rewards were introduced to promote balance maintenance, energy minimization, and smooth transitions between locomotion modes such as walking, climbing, and tilting. The specific reward terms are listed in Table~\ref{tab:rewards}

\begin{table*}[ht]
    \centering
    \begin{tabular}{c|c|c}
        \toprule
        reward & expression & scale \\
        \midrule
        tracking goal vel & $\min \left( \frac{\vec{v} \cdot \hat{\vec{t}}}{v_{cmd} + 10^{-5}}, \frac{v_{cmd}}{v_{cmd} + 10^{-5}} \right), \quad \text{where } \hat{\vec{t}} = \frac{\vec{t}}{||\vec{t}|| + 10^{-5}}$ & 1.5 \\
        tracking yaw vel & $\exp \left( -|\omega_z - \omega_{cmd}| \right)$ & 1.  \\
        tracking pitch & $\exp \left( -3 |p_{cmd} - p| \right)$ & 1.5\\
        lin vel z walking & $v_z^2$ & -9.0\\  
        ang vel xy & $\sum \omega_{xy}^2$ & -0.05\\
        dof acc & $\sum \left( \frac{\dot{q}_{t+1} - \dot{q}_t}{\Delta t} \right)^2$ &-2.5e-7\\
        collision & $\sum \mathbf{1} \left( ||f_{\text{contact}}|| > 0.1 \right)$ & -5.  \\
        action rate & $||\mathbf{a}_{t+1} - \mathbf{a}_t||$ &  -0.1\\
        delta torques & $\sum (\tau_{t+1} - \tau_t)^2$ & -1.0e-7\\
        torques & $\sum \tau^2$ &-0.00001\\
        hip pos & $\sum (q_{\text{hip}} - q_{\text{hip, default}})^2$ &-1\\
        dof error & $\sum (q - q_{\text{default}})^2$ &-0.2\\
        feet stumble & $\mathbf{1} \left( ||f_{\text{contact, xy}}|| > 4 \cdot |f_{\text{contact, z}}| \right)$ & -5\\
        feet edge & $\sum \mathbf{1}(\text{feet at edge})$ &-1\\
        feet drag & $\sum \left( \mathbf{1}(\text{contact}) \cdot ||v_{\text{xy}}^{\text{feet}}|| \right)$ &-0.1\\
        energy & $||\tau \cdot \dot{q}||$ &-1e-3\\
        \bottomrule
    \end{tabular}
\caption{\textbf{Reward terms}}
\label{tab:rewards}
\end{table*}

\subsubsection{Details of Deployment}
Depth images were captured using a Realsense D435 camera connected to the Nvidia Jetson Orin via a USB 3.0 interface. We applied hole-filling filters, spatial filters, and temporal filters, followed by resizing and normalization—mirroring the process used in simulation. The depth encoder network operates at 10 Hz with a fixed delay and communicates with the main process via UDP. The main process executes the distilled policy at 50 Hz, while proprioceptive data is obtained at 500 Hz through Cyclone DDS. Computed joint angles and PD parameters are transmitted to the Unitree low-level controller via ROS 2 messages, where motor torques are calculated using the internal PD controller.

\subsection{Details of Zero-shot Deployment Using VLMs}
As the robot approaches the target object, a transition to a precise grasping policy is triggered, allowing for more accurate command following. This transition is governed by the overhead camera and occurs when the robot is within 1 meter of the target object and oriented within 30 degrees of it. After successful grasping, the policy switches again once the robot is aligned within 30 degrees of the termination point.